\documentclass[letterpaper]{article}

\usepackage{paperstyle}
\usepackage{times}
\usepackage{helvet}

\usepackage[hyphens]{url}
\usepackage{graphicx}
\usepackage{natbib}
\usepackage{caption}
\usepackage{longtable}
\usepackage{multirow}
\usepackage{amssymb}
\usepackage{makecell}
\usepackage{placeins}
\usepackage{amsmath}

\title{Safe by Benchmark, Unsafe in Practice: \\ Lessons from 1331 Child-Related AI Incidents}

\author{Junfeng Jiao\textsuperscript{1}, Kevin Chen\textsuperscript{1},
Shrestha Mishra\textsuperscript{1,*}, Abhejay Murali\textsuperscript{1},
David Atkinson\textsuperscript{2}, Shriti Raj\textsuperscript{3}, and
Saleh Afroogh\textsuperscript{1,$\dagger$}}

\affiliations{
\textsuperscript{1}The University of Texas at Austin, USA\\
\textsuperscript{2}Allen Institute for AI, USA\\
\textsuperscript{3}Stanford University, USA
}

\begin{document}

\maketitle

\begingroup
\renewcommand{\thefootnote}{\fnsymbol{footnote}}
\footnotetext[2]{Lead author: saleh.afroogh@utexas.edu}
\footnotetext[1]{Corresponding author: shresthamishra@utexas.edu}
\endgroup

\begin{abstract}
AI systems are increasingly present in children's everyday environments—education, entertainment, social networks, and toys. Their safety is typically assessed at the model level, through benchmarks, red-teaming, and content filtering. However, most AI harm taxonomies lack child-specific evidence and under-specify developmental harm and the stakeholder asymmetries unique to minors. Building on prior taxonomies, we construct a child-centered harm taxonomy mapped to sociotechnical failure modes. We curate and code 1331 publicly reported child-related AI incidents, annotating internal and external harm types and the failure mode implicated in each incident. Our analysis reveals which harms are most prevalent in public reporting, how individual-level and population-level harms relate, and which failures are most often coded alongside harm. We find that harm concentrates in representational, dignity-related, and psychological categories, and co-occurs most often with deployment-context, governance, and misuse failures rather than model defects. Harm to an individual child is also frequently linked to population-level harm, reaching children who never interacted with the system. These findings suggest that model-level evaluations, while necessary, cannot by themselves anticipate the harm pathways most often documented for children. We conclude with actionable implications emphasizing mitigations beyond model-level safeguards such as content filtering: age assurance, consent verification for depiction, safety-by-design UX constraints, incident disclosure norms, and auditing regimes targeting developmental risk. Privacy-minimized annotations, a codebook, and documentation are available at \url{https://github.com/The-Responsible-AI-Initiative/KidAID}.
\end{abstract}

\section{Introduction}

AI systems are rapidly moving from general-purpose tools into child-facing settings: tutoring chatbots, creative assistants, ``AI friends,'' voice assistants, social platforms, and networked toys.
This expansion creates tangible benefits in personalized learning support, accessibility, entertainment, but also introduces a distinct harm landscape for minors.
Children differ from adults in cognitive development, risk perception, susceptibility to persuasion, and dependence on guardians and institutions \cite{unicef2021,weidinger2023}.
As a result, harms that are rare or low-severity in adult contexts can become high-severity in child contexts, especially when systems shape emotions, relationships, or identity formation.
These differences map onto specific affordances and failure modes.
Susceptibility to persuasion meets anthropomorphic conversational design, the setting for overreliance failures and dependency harms.
Limited capacity to consent meets generative depiction, the setting for sexualization harms.
Deference to perceived authority meets systems framed as tutors or experts, where unsafe advice is more likely to be followed than it would be by an adult.
Dependence on caregivers and institutions means the harmed child is frequently not the party who accepted terms, configured the system, or can seek recourse.
This mapping structures the taxonomy that follows.

The public debate on the risk of AI often divides between long-range catastrophic concerns and current harm \cite{bostrom2014,hendrycks2023,bender2021,crawford2021,birhane2022}.
For children, present-day harms are already prominent and measurable: exposure to sexual content, grooming or manipulation, encouragement to self-harm, parasocial dependency, privacy leakage, and coercive monetization patterns  \cite{commonsense2024,livingstone2021,iwf2024,unicef2021}.
Despite growing attention, child AI safety research remains fragmented between (i) parental concern surveys, (ii) lab-style safety evaluations, (iii) platform policy documents, and (iv) isolated high-profile incidents.
This fragmentation makes it difficult to prioritize mitigations because stakeholders lack empirically grounded answers to two operational questions.
\textbf{RQ1 (What):} \textit{What is the nature and prevalence of child-specific AI harms observed in practice, at both the individual and the population level?}
\textbf{RQ2 (How):} \textit{Which sociotechnical failure modes are most often coded alongside child AI harms, and at what level of the sociotechnical system do they occur?} \cite{anderljung2023,eu2024,nist2024}.

Existing taxonomies of AI harms and failures provide critical vocabulary but typically treat ``psychological harm'' as a broad bucket, under-represent child-specific threats (grooming, sexualization, coercive roleplay), and lack datasets mapping harms to failure modes and stakeholders in real incidents.

To address these gaps, we contributed a child-centered incident dataset and a taxonomy grounded in empirical evidence.
Our approach mirrors incident-analysis work in other safety-critical fields: we treat each publicly reported child-related AI event as a case, annotate what happened at both the individual and population level and how the sociotechnical system failed, then quantify prevalence patterns and map harm-to-failure pathways to highlight leverage points for mitigation.

\section{Related Work}

The AI systems used around children are at the intersection of technical risk, governance, and developmental vulnerability.
Recent work on generative AI has emphasized that harm is not produced only by the capability of the model, but also by human interaction and the broader context of deployment \cite{weidinger2023}.
This sociotechnical framework is especially important for children with AI, where harms often emerge through repeated interaction, authority signals, or emotional dependence.

Incident-based analysis offers a complementary empirical base.
The AI Incident Database was created to catalog real-world failures and support organizational learning \cite{mcgregor2021}, and recent incident-mapping work on generative AI extends this approach by jointly analyzing \textbf{what} harms occur, \textbf{who} is harmed, and \textbf{how} the underlying failures arise in practice \cite{li2025}.
Our study follows this incident-analytic logic but applies it to child-related AI harms.
We are especially inspired by \citet{li2025}, which demonstrates how incident-based analysis can operationalize accountability by jointly modeling what happened, who was harmed, and how harms emerged.
Following this ethos, we curate child-related incidents from AIAAIC, AIID, and OECD.AI \cite{aiaaic2025,mcgregor2021,oecd2025}.

UNICEF's policy guidance emphasizes that AI governance should reflect children's developmental needs, evolving capacities, dependency relations, and rights-based protections \cite{unicef2021}.
Governance frameworks such as the NIST Generative AI Profile and the OECD AI principles stress lifecycle risk management, transparency, and human-centered safeguards \cite{nist2024,oecd2019}.

\subsection{Taxonomies of AI Risks and Harms}

Risk and harm taxonomies standardize the way stakeholders describe negative outcomes, enabling measurement, auditing, and governance \cite{shelby2023,weidinger2023}.
Prior work has proposed broad harm categories spanning physical safety, privacy, civil liberties, misinformation, economic harm, psychological harm, and societal impacts, with generative AI-specific taxonomies further emphasizing modality-driven risks (e.g., deepfake impersonation) and systemic impacts.
Governance-facing frameworks (e.g., the NIST AI Risk Management Framework) encourage practitioners to map risks to concrete controls across design, development, deployment, and monitoring.

Child safety requires finer granularity than general adult taxonomies.
In child contexts, hazards like sexual content are not simply ``hazardous content exposure''; they include sexualization of minors, grooming, coercive roleplay, and depiction/consent violations whose severity depends on developmental stage and asymmetric power dynamics.
Likewise, ``psychological harm'' must represent developmentally specific mechanisms (e.g. persuasion susceptibility, attachment formation, authority compliance, and dependence).
Our taxonomy extends general risk vocabularies by adding child-specific harm types and by separating individual-level harms from the population-level harms that reach children who never interacted with the system (e.g., non-interacting child victims depicted in synthetic media).

\subsection{Sociotechnical Failure Modes and Safety Engineering Views}

We use three terms that carry much of the paper's conceptual weight. \textit{Failure mode} is inherited from safety engineering, where it denotes the specific manner in which a system departs from intended function \cite{leveson2011}. \textit{Sociotechnical} is adapted from prior harm taxonomies, which argue that harm arises jointly from model capability, human interaction, and deployment context rather than from model outputs alone \cite{weidinger2023,shelby2023}. \textit{Harm-to-failure pathway} is our own term, defined operationally as the co-occurrence, within a single incident, of a harm code and its primary failure-mode code; it asserts documented association, not causal mechanism. Safety engineering concepts motivate mapping harms to the proximal mechanisms associated with them.
Work on AI failures distinguishes technical failures (model hallucination, robustness gaps, security vulnerabilities) from sociotechnical issues (misaligned incentives, misuse, weak disclosure practices, governance failures) \cite{weidinger2023}.
Governance frameworks urge practitioners to treat AI as a lifecycle-managed sociotechnical system where failures can be introduced during data collection, model development, product design, deployment configuration, and post-deployment operations.

Child harms frequently manifest as multi-factor pathways: weak or bypassable age gates, engagement-optimized recommender incentives, persuasive conversational UX patterns, and missing crisis escalation procedures can jointly produce severe outcomes.
These compounded pathways motivate explicit failure-mode coding, including misuse, robustness failures, privacy leakage, inadequate oversight, and governance noncompliance, so mitigations can be targeted to the stage and mechanism that enabled harm \cite{leveson2011}.

\subsection{AI Incidents as Empirical Data}

Incident repositories and public reporting have become a pragmatic empirical base for studying AI harms, enabling quantification of harm prevalence, taxonomy validation, and identification of accountability gaps.
The OECD has advanced a \textit{common reporting framework} for AI incidents to standardize cross-case comparison and governance learning \cite{oecd2025}.
The incident approach has well-known limitations in selection effects, what gets reported, missing technical details, inconsistent causal attribution, and uneven geographic coverage, but it anchors harm taxonomies in observed failures rather than hypothetical possibilities.
For child safety in particular, consolidating dispersed incidents (journalism, lawsuits, regulatory actions, school disclosures, platform enforcement reports) into a structured dataset enables reproducible measurement of prevalence patterns and harm--failure pathways.

\subsection{``How--What--Who'' Mapping for Accountability}

A central contribution of recent incident-driven analyses is to show that accountability requires simultaneously modeling: \textbf{(i) what harm occurred, (ii) who was harmed, and (iii) how the harm arose} \cite{li2025}.
This framing is especially salient for child harms, where the harmed party is often not the account holder or direct user, such as minors depicted in synthetic content or children exposed via recommendation pathways.
Standard end-user-centric governance can therefore mis-assign responsibility and under-incentivize upstream controls.
Our work extends this mapping by specializing it for child stakeholders and developmental harm mechanisms.
We depart from prior incident-analytic work \cite{li2025} in three ways.
First, \textit{dual-level harm coding}: each incident receives both an internal (individual-level) and an external (population-level) harm code, so the link between harm to a child and its wider circulation can be measured rather than asserted (Appendix~A).
Second, \textit{taxonomic granularity}: below adult-general categories we code child-specific harm types, including sexualization, sexual exploitation, addiction/compulsive use, over-reliance/dependency, and self-harm.
Third, \textit{scope}: prior child--AI work is survey-based, evaluation-based, or single-case; to our knowledge this is the first corpus-scale incident study restricted to minors.

\section{Methodology}

We perform a systematic, post-hoc incident analysis of publicly reported AI-related events in which a minor (under 18) plausibly experienced harm or a credible worst-case risk of harm.
We define an incident as a discrete event or episode that became public through credible reporting (e.g., regulator actions, lawsuits, vulnerability disclosures, or investigative journalism).

We curate candidate incidents from public incident repositories, regulatory announcements, court filings, cybersecurity/vulnerability disclosures, and credible media reports.
We include an item if (i) AI is materially involved (generation, recommendation, automated decision, or AI-mediated interaction); (ii) at least one affected stakeholder is a minor; and (iii) the report contains enough detail to code a harm category and a failure mode.
We exclude purely speculative claims without an observed event, cases where child involvement is not plausible, and duplicates.
The final corpus contains 1331 incidents: 1143 from the OECD.AI incident monitor, 141 from AIAAIC, and 47 from the AI Incident Database.
The \textit{Occurred} field is taken from the source repository and follows its dating convention: OECD.AI records a full date, AIAAIC and AIID record a year, and one AIAAIC record gives a multi-year range.
Temporal analyses therefore use year of occurrence.

Each incident is manually annotated using a structured coding schema.
We code (1) child-specific harms (one internal (individual-level) harm and one external (population- or institutional-level) harm per incident; refining general taxonomies into developmentally salient categories such as sexual harms, self-harm/mental-health escalation, manipulation/dependency, privacy/security leakage, misinformation, harassment/extortion, physical safety harms, and rights/procedural harms); and (2) sociotechnical failure modes, defined as the proximal mechanisms that are technical, organizational, governance, or use-related, through which an AI sociotechnical system is implicated in harm \cite{weidinger2023,shelby2023}.
Each incident is coded with one primary failure mode: the proximal mechanism most directly implicated in the reported harm. Where reports document multiple contributing failures, we record the most proximal \cite{leveson2011,raji2022}.
Failure modes are organized by lifecycle stage: design, development/evaluation, release and use.

We develop the codebook iteratively (released alongside the dataset and reproduced in Appendix~B): we seed categories from established taxonomies, introduce child-specific refinements, and pilot-code an initial subset.
 The remaining incidents are then independently annotated by two members of the research team, and disagreements are resolved through consensus coding \cite{hill2005,saldana2021}.
To assess coding reliability, we calculated inter-rater agreement on a
randomly sampled subset of 150 incidents (11.3\% of the dataset) before
consensus (see Appendix section ``Inter-Rater Reliability Calculation''). For categorical variables (harm categories and failure modes),
we computed Krippendorff's $\alpha$: harm categories
$\alpha = 0.73$ and failure modes $\alpha = 0.67$.
For binary worst-case flags, Cohen's
$\kappa = 0.58$. These values indicate acceptable agreement for harm categories ($\alpha = 0.73$) and only moderate agreement for failure modes ($\alpha = 0.67$) and worst-case flags ($\kappa = 0.58$).
Krippendorff's conventional threshold for drawing firm conclusions is $\alpha \geq 0.80$, with $0.67 \leq \alpha < 0.80$ supporting tentative conclusions only \cite{krippendorff2018}.
We therefore treat all failure-mode-dependent findings as provisional and interpret worst-case flags as indicative rather than definitive.
Agreement was lowest where category boundaries are inherently contested --- distinguishing contextual norm violation from governance noncompliance, and improper from malicious use --- a limitation we return to below.

When reports cannot confirm whether harm has fully materialized, we record a plausible worst-case flag and treat these cases conservatively.
From the coded corpus we compute three families of descriptive statistics: frequency distributions over harm categories, harm types, and failure modes; co-occurrence counts between harm categories and failure modes, and between internal and external harms, constructed by tallying incidents in which both codes appear; and annual counts of the most frequent failure modes.
All reported quantities are counts of coded incidents. We do not fit models or test hypotheses.

This approach has notable limitations.
Because incident narratives are observational and variably detailed, we avoid causal claims and report associations that identify high-priority harm pathways.
In 112 of the 1331 incidents (8.4\%), the available reporting did not support assignment of an internal harm category (two of these also lack an external harm code); these incidents are retained in the corpus and in all failure-mode and external-harm analyses, but are excluded from internal harm category counts. Key validity threats include reporting bias, incomplete technical details, age ambiguity, platform-level non-independence, and limited causal identifiability; we mitigate these through uncertainty flags and conservative worst-case labeling \cite{krippendorff2018,hruschka2004,rosenbaum2002,shadish2002,turri2023,pittaras2022}.

% FIGURE 1 — coding schema table

\begin{figure*}[t]
    \centering
    \includegraphics[width=\textwidth]{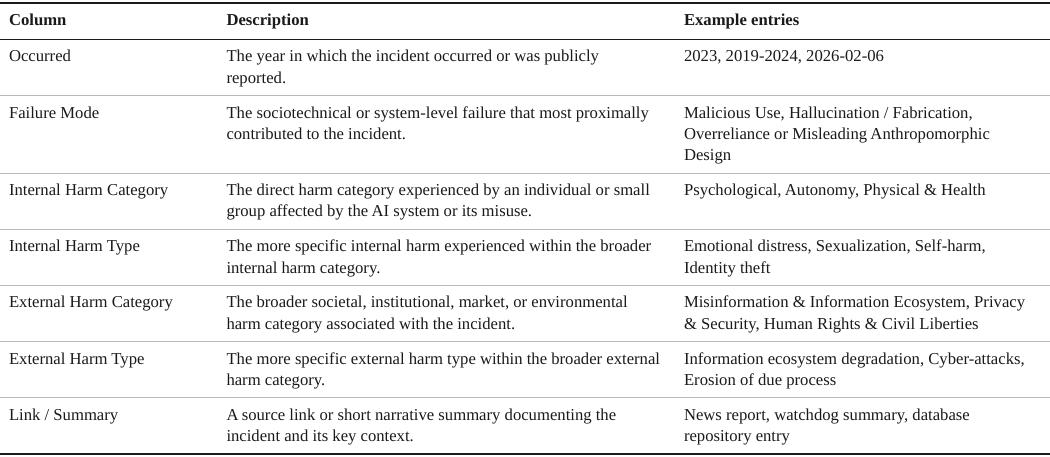}
    \caption{Overview of Dataset Coding Schema.}
    \label{fig:coding_schema}
\end{figure*}

The schema is hierarchical: high-level harm categories were seeded from prior taxonomies \cite{shelby2023,weidinger2023}, and harm types are sub-codes that aggregate into these categories.
The taxonomy is therefore neither purely deductive nor purely inductive.
Categories were seeded deductively from adult-general taxonomies and largely retain their structure; child-specific refinement happens mainly at the harm-type level, where pilot coding surfaced cases that adult-general types could not hold without loss --- principally sexualization, sexual exploitation, addiction/compulsive use, and over-reliance/dependency.
We also code internal and external harm separately for every incident.

\section{Applying the Schema: Three Worked Cases}

To make the coding schema concrete, we trace three incidents from report to code to implicated mitigation.
The cases illustrate three recurring deployment patterns in our corpus: generative depiction, companion interaction, and institutional deployment.
Each incident carries one primary failure mode; contributing failures documented in reporting are noted here but are not counted.

\subsection{Case 1: Synthetic Sexual Depiction (Almendralejo, 2023)}
\textit{What happened.} Students at a secondary school in Almendralejo, Spain used a commercial ``nudification'' application to generate synthetic nude images of female classmates from ordinary clothed photographs; the images then circulated through peer messaging groups \cite{reuters2023}.
\textit{Harm codes.} Internal: Representation, Dignity \& Social Identity / sexualization --- coded here rather than Psychological because the harm is constituted by how the child is depicted to others, not by the child's own interaction with the system. External: Representation \& Toxicity (Population-Level) / toxic content.
\textit{Primary failure mode.} Malicious use. Contributing: a general-purpose image tool available without restriction on depicting identifiable minors.
\textit{Implicated mitigation.} The depicted girls never used the application, so controls conditioned on the user cannot reach this case. It implicates depiction-consent verification and restriction at generation (second roadmap priority).

\subsection{Case 2: Companion Chatbot (Character.AI, 2024)}
\textit{What happened.} A fourteen-year-old developed a sustained relationship with a persona-based conversational agent over several months, culminating in his death by suicide \cite{roose2024}.
\textit{Harm codes.} Internal: Physical \& Health / loss of life. External: Physical (Societal/Public Safety) / damage to public health. The harm is coded by its reported outcome; the psychological pathway that preceded it is captured by the failure mode.
\textit{Primary failure mode.} Overreliance or misleading anthropomorphic design. Contributing: no crisis routing on self-harm signals, and emotionally intimate framing available to a minor.
\textit{Implicated mitigation.} No single turn need cross a safety threshold for the cumulative interaction to end in harm, so per-output content evaluation cannot reach this case. It implicates session-level safety evaluation, guardian visibility, and mandatory escalation (first roadmap priority).

\subsection{Case 3: Institutional Deployment (School Weapons Detection, 2025)}
\textit{What happened.} An AI weapons-detection system at a Baltimore-area high school flagged a student's bag of chips as a firearm; armed police responded and handcuffed the student \cite{aiaaic2025}.
\textit{Harm codes.} Internal: Autonomy / loss of rights and freedoms --- coded as Autonomy rather than Physical \& Health because no bodily injury occurred and the operative harm is deprivation of liberty. External: Human Rights \& Civil Liberties / erosion of due process.
\textit{Primary failure mode.} General robustness and reliability failure. Contributing: no human confirmation step before an armed response.
\textit{Implicated mitigation.} The model error is ordinary; the severity comes from the escalation procedure around it. It implicates child-specific pre-deployment assessment with institutional consequence modeling (third roadmap priority).

\subsection{What the Cases Show}
The three cases differ in where the failure enters and who holds the control point.
In Case 1 the decisive control lies upstream, in what the generative system will produce; in Case 2 it lies in interaction design and is invisible to per-output evaluation; in Case 3 it lies with the deploying institution.
No single intervention point covers all three, which is why the roadmap that follows is layered.

\section{Qualitative Findings and Resultant Themes}

We report the child-specific harms and sociotechnical failure modes surfaced in our coded incident data set (1331 publicly reported child-related AI incidents), their prevalence, how individual-level and population-level harms relate, and how harms and failures relate to each other.
Each incident carries one primary failure mode, so failure-mode counts sum to $N$. Harm counts are reported separately for internal and external harms, each coded once per incident.
Our primary contributions here are (i) an empirically grounded child-centered harm taxonomy, (ii) a lifecycle-staged taxonomy of child-relevant sociotechnical failure modes, and (iii) relational mappings between the harms that occur and the failures they co-occur with.
Throughout, \textit{prevalence} denotes frequency within our corpus of publicly reported incidents, not incidence in the underlying population of child--AI interactions.
The two diverge systematically: incident reporting favors acute, legally actionable, and media-legible events over slow-accumulating, private, or normalized harms.
All distributional claims below should be read under this qualification.
\subsection{What Child-Specific AI Harms Surface in Publicly Reported Incidents?}

Our taxonomy of child-related AI harms is organized into eleven high-level categories, each comprising child-specific sub-harms that refine adult-general taxonomies to account for developmental vulnerability, constrained capacity for consent, and caregiver-mediated recourse.

Across all coded incidents,\footnote{Internal harms are individual-level harms experienced directly by a child (e.g., sexualization, emotional distress, privacy loss). External harms are population- or society-level harms that scale beyond the directly affected child (e.g., toxic content circulation, public-safety risks, reputational damage). See Appendix~B for full definitions and representative examples.} the most common internal harm categories are Representation, Dignity and Social Identity ($n = 475$) and Psychological harms ($n = 409$), followed by Autonomy ($n = 135$) and Physical \& Health harms ($n = 85$).
Smaller but still meaningful categories include Privacy, Identity \& Information Security (Individual) ($n = 75$), Political \& Economic ($n = 14$), Financial, Property \& Material ($n = 9$), Intellectual Property \& Creative Labor ($n = 8$), Misinformation \& Information Ecosystem ($n = 6$), Human Rights \& Civil Liberties ($n = 2$), and Reputational ($n = 1$).

This distribution suggests that child-related AI harm is not primarily concentrated in narrow technical-error categories (such as privacy/security leakage from model misconfiguration or hallucinated factual errors).
Instead, it is heavily shaped by harms to dignity, identity, emotional well-being, and autonomy.
Within Representation, Dignity \& Social Identity, sexualization ($n = 316$) is by far the most prevalent subtype, followed by sexual exploitation ($n = 131$) and exposure to toxic content ($n = 17$).
This ranking should be read as a statement about what surfaces in public reporting rather than as evidence that sexualization outweighs dependency formation, subtle persuasion, or chronic emotional dysregulation --- harms our own framework identifies as developmentally significant but which are structurally under-reported because they accumulate privately and rarely generate a discrete reportable event.
Within the Psychological category, the most frequent subtypes are addiction/compulsive use ($n = 79$), self-harm ($n = 69$), health deterioration ($n = 66$), and toxic content exposure ($n = 62$).

\begin{figure*}[t]
    \centering
    \includegraphics[width=\textwidth]{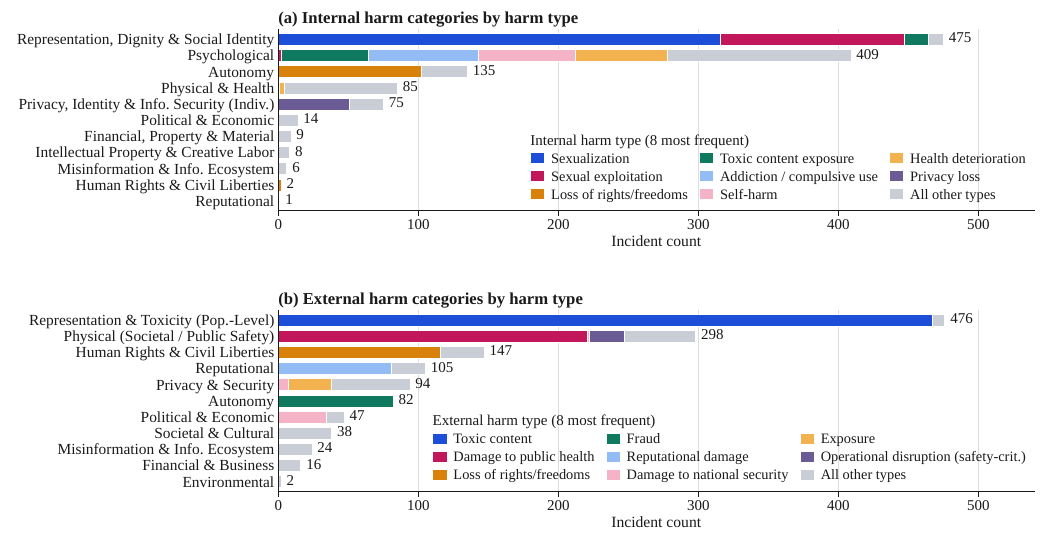}
    \caption{Internal (a) and external (b) harm categories by harm type; the eight most frequent types are shown individually.}
    \label{fig:harm_by_type}
\end{figure*}

The category-to-type mapping is visualized in Appendix~A, Figure~\ref{fig:internal_cooccurrence} (internal), and in Figure~\ref{fig:harm_by_type}(b) (external).

The external harm structure shows a parallel pattern at the societal level.
The most common categories of external harm are Representation \& Toxicity (Population-Level) ($n = 476$), Physical (Societal/Public Safety) ($n = 298$), Human Rights \& Civil Liberties ($n = 147$), Reputational ($n = 105$), and Privacy \& Security ($n = 94$). Less frequent external categories are Autonomy ($n = 82$), Political \& Economic ($n = 47$), Societal \& Cultural ($n = 38$), Misinformation \& Information Ecosystem ($n = 24$), Financial \& Business ($n = 16$), and Environmental ($n = 2$).
This dual structure reflects an ecological model of harm in which individual- and population-level harms are connected through circulation and re-injury \cite{bronfenbrenner1979,livingstone2021}.

% FIGURE 3 — Taxonomy of Internal Harms

\begin{figure*}[t]
    \centering
    \includegraphics[width=\textwidth]{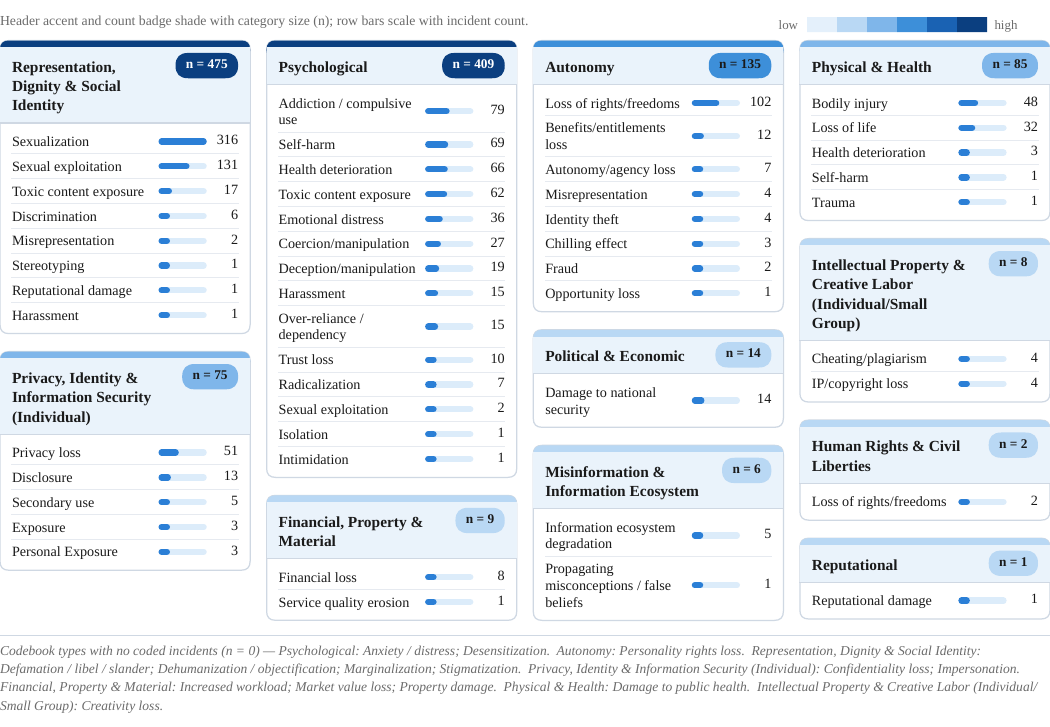}
    \caption{Taxonomy of Internal Generative AI Harms (ordered by prevalence).}
    \label{fig:internal_harms}
\end{figure*}

\subsection{How Do Child AI Harms Surface? Prevalence of Sociotechnical Failure Modes}

The failure-mode distribution shows that child-related AI harms are most often coded with Human, Organizational, and Contextual Deployment Failures ($n = 779$) --- failures in how AI systems are situated, governed, and used in real social contexts.
Within this group, violation of contextual norms ($n = 398$) is the single largest failure mode in the entire data set, followed by Governance/Regulatory Noncompliance ($n = 227$) and Overreliance or Misleading Anthropomorphic Design ($n = 146$).
Many harms emerge not from bad model outputs but from deployment in contexts where safety and age-appropriateness expectations are violated.

For example, Snap's 2023 deployment of ``My AI'' inside an app with a large minor userbase, in which the system discussed age-inappropriate topics with accounts registered to 13-year-olds, was primarily a contextual norm violation rather than a model defect \cite{fowler2023}.
Similarly, the 2023 FTC action against Amazon for retaining children's Alexa voice recordings after parental deletion requests illustrates how governance noncompliance compounds privacy leakage \cite{ftc2023}.

The second largest group is Misuse/Abuse ($n = 297$), dominated by Malicious Use ($n = 228$).
Model Behavior/Technical Failure is smaller overall ($n = 196$), where General Robustness/Reliability Failure ($n = 128$) far exceeds Hallucination/Fabrication ($n = 47$), indicating that reliability under real deployment conditions matters more than hallucination alone.
Privacy, Data Governance and Information Handling ($n = 59$) is driven almost entirely by Privacy or Confidentiality Leakage.
The four failure-mode families operate at different levels, and the level determines which actor can intervene.
Model Behavior/Technical failures (robustness, hallucination, data quality) are properties of the model or pipeline and fall to developers.
Privacy and data-governance failures concern how children's data are collected and handled, and fall to developers and deployers jointly.
Human, Organizational, and Contextual Deployment failures (contextual norms violation, governance noncompliance, overreliance or misleading anthropomorphic design, inadequate oversight) are properties of the product, the deploying institution, and the rule environment.
Misuse/Abuse failures (malicious, undisclosed, or out-of-scope use) arise at the point of use and fall to platforms and the controls they place on users.
Because each incident carries one primary code, these counts locate the most proximal failure, not every failure in the chain.

% FIGURE 4 — Taxonomy of Sociotechnical Failure Modes

\begin{figure*}[t]
    \centering
    \includegraphics[width=\textwidth]{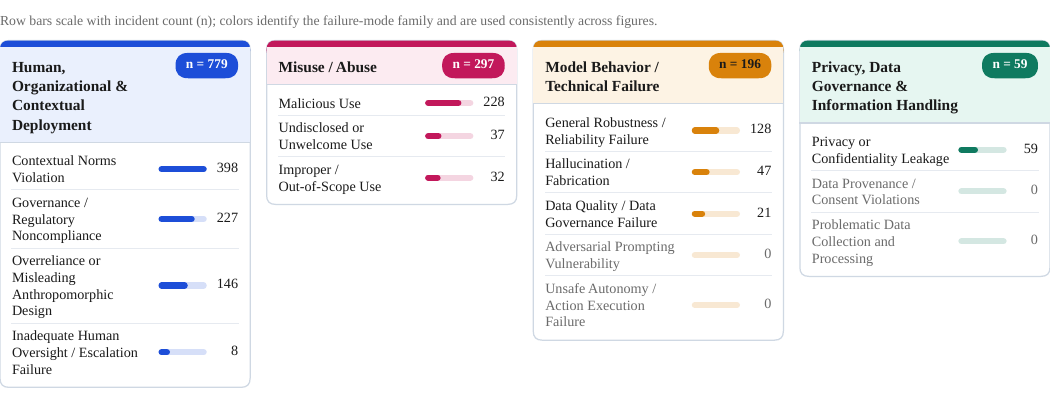}
    \caption{Taxonomy of Sociotechnical Failure Modes (bar length indicates prevalence; colors identify the failure-mode family).}
    \label{fig:failure_modes}
\end{figure*}

\subsection{Which Failure Modes Are Associated with Each Internal Harm Category?}

The largest internal harm categories, that is Representation, Dignity and Social Identity and Psychological harms align with the prominence of Contextual Norms Violation, Malicious Use, and Overreliance or Misleading Anthropomorphic Design.
Harms to children are thus most often documented alongside socially inappropriate use, deliberate exploitation of tools, or interaction design that encourages emotional or social over-trust.
The prevalence of sexualization, exploitation, and toxic exposure is consistent with a landscape of failure where malicious or norm-violating use plays a central role.
Design patterns associated with dependency or misleading social attachment are not peripheral: overreliance or misleading anthropomorphic design is the primary failure mode in 146 incidents, 49 of them with a Psychological internal harm.
The Autonomy and Physical \& Health categories, while smaller, are associated with improper or out-of-scope use, contextual norm violation, and overreliance on system outputs confirming that internal harms are often multifactorial.

\subsection{Which Failure Modes Are Associated with External Harm Categories?}

The single largest external category, Representation \& Toxicity (Population-Level), is dominated by toxic content ($n = 467$), with the remainder distributed across discrimination at population or systemic scale ($n = 3$), stereotyping ($n = 2$), group-level stigmatisation ($n = 2$), and group-level marginalisation ($n = 2$).
For example, AI-generated nude images of a single classmate frequently spread across an entire school cohort through social networks, transforming an individual harm into peer-network exposure and reputational damage that affects dozens of uninvolved minors \cite{singern2023,reuters2023}.
This aligns with the dominance of Contextual Norms Violation and Malicious Use, since population-level harm depends on circulation, amplification, and deployment context.

Physical (Societal/Public Safety) ($n = 298$) and Human Rights \& Civil Liberties ($n = 147$) indicate that child-related AI incidents sometimes involve broader rights protections and public safety consequences.
Three crosscutting themes emerge: (1) child-related AI harms are strongly sociotechnical, and the largest failure families are deployment, governance, and misuse-related; (2) the most prevalent harms concentrate in psychological, representational, and dignity-related categories; and (3) many harms are multi-layered, with direct individual harms and broader societal harms co-emerging. This supports layered governance and safety approaches \cite{leveson2011,weidinger2023,li2025}.

% FIGURE 5 — Taxonomy of External Harms (vertical)

\begin{figure*}[t]
    \centering
    \includegraphics[width=\textwidth]{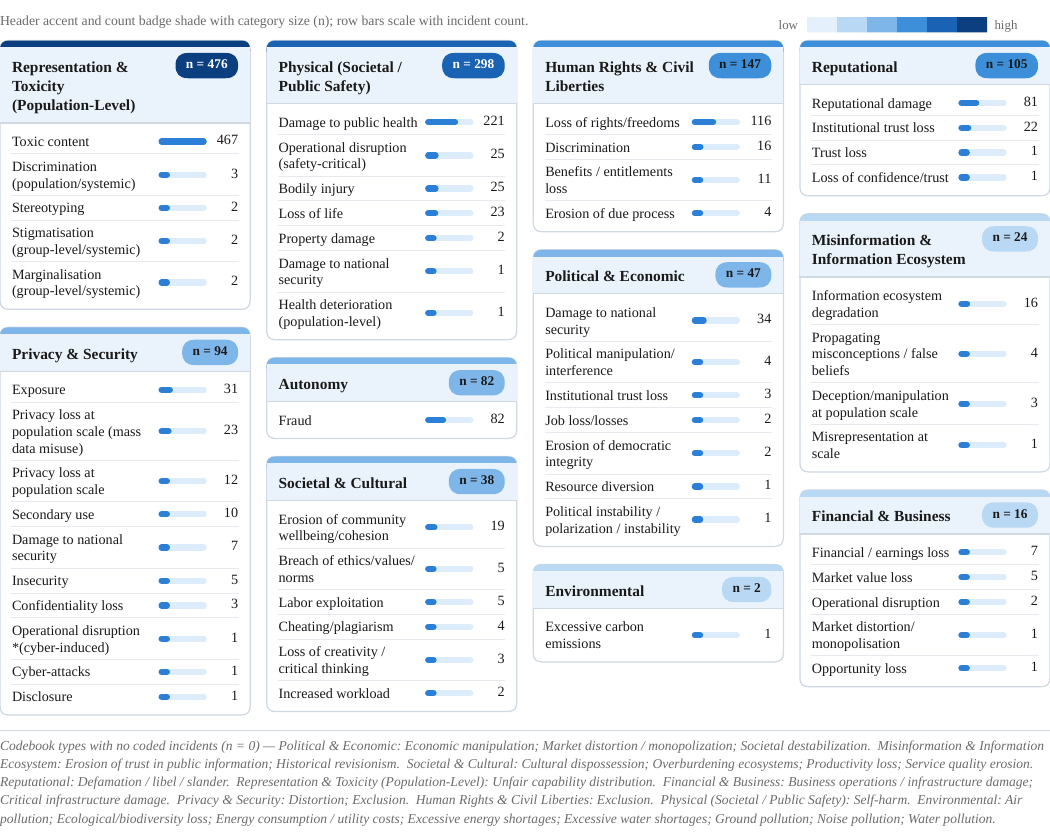}
    \caption{Taxonomy of External Generative AI Harms (darker colors indicate higher prevalence).}
    \label{fig:external_harms}
\end{figure*}

\section{Quantitative Findings and Resultant Patterns}

Our quantitative analysis draws on child-related AI incidents curated from the AIAAIC, AIID, and OECD.AI incident monitor.
These repositories offer a pragmatic empirical basis while reflecting known limitations: reporting bias, uneven documentation quality, and stronger visibility for acute events than cumulative harms \cite{abercrombie2024,mcgregor2021,oecd2025,turri2023}.
We treat each incident as an observational case coded with structured attributes spanning failure mode and internal and external harm category \cite{hutiri2024,raji2022,velazquez2024}.

\subsection{How Do Generative AI Harms Surface in Publicly Reported Incidents?}

The temporal trend in our data set suggests that child-related AI harms surface through a progressively broader sociotechnical failure landscape rather than a single dominant technical defect.
Incident counts were low and flat through the mid-2010s, rose sharply beginning around 2020, and especially accelerated after 2022.
Contextual Norms Violation becomes the most prominent failure mode in the later years, peaking dramatically in 2024--2025, reflecting systems being introduced into settings where expectations about safety, age-appropriateness, supervision, and consent are violated.
The sharp increase in malicious use beginning in the early 2020s (accelerating in 2023--2025) is consistent with the growing visibility of incidents involving deepfake sexual abuse material, grooming facilitation, and exploitative impersonation.

Publicly reported child-related incidents are especially likely to surface where harms are highly visible, institutionally legible, or socially alarming, which helps explain why the latter period is dominated by malicious exploitation and contextual norm breakdown rather than slower cumulative harms such as dependency formation or subtle persuasion \cite{pittaras2022,raji2022,velazquez2024}.

% FIGURE 6 — Yearly trend of top failure modes

\begin{figure*}[t]
    \centering
    \includegraphics[width=\textwidth]{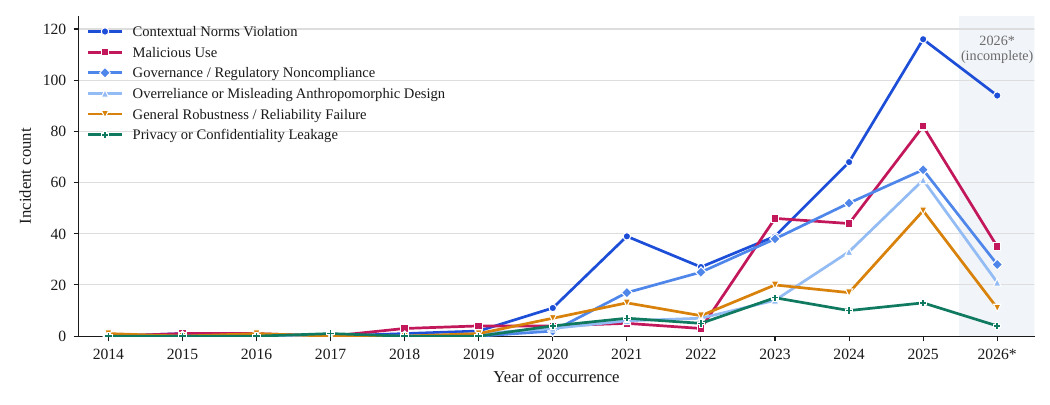}
    \caption{Most prevalent failure mode per year, 2014--2026$^{*}$ ($^{*}$incomplete year).}
    \label{fig:temporal_trends}
\end{figure*}

\subsection{Which Failure Modes Are Associated with Each Type of Harm?}

\begin{figure*}[t]
    \centering
    \includegraphics[width=\textwidth]{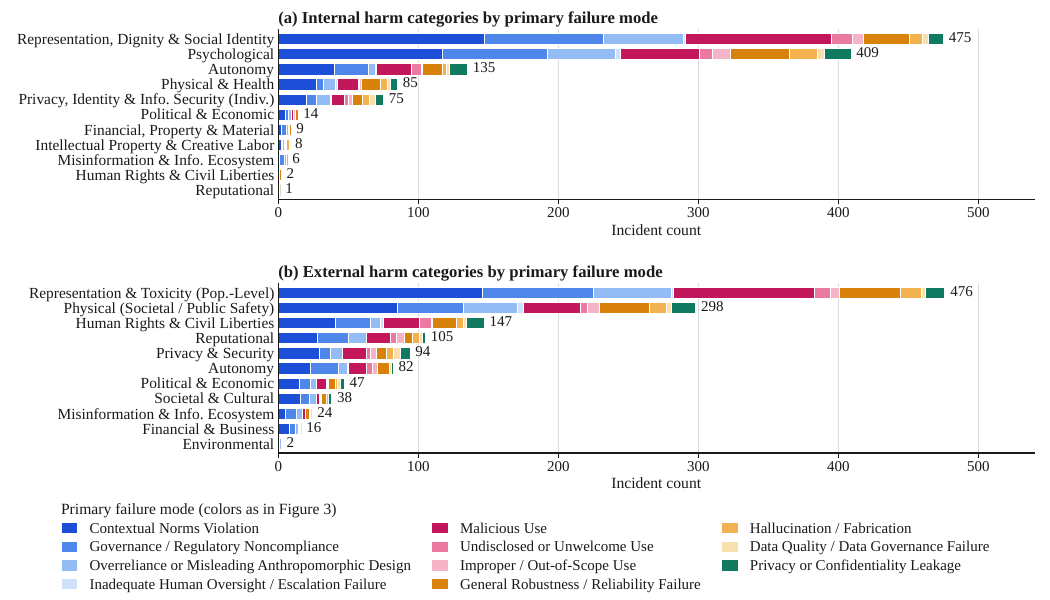}
    \caption{Internal (a) and external (b) harm categories by primary failure mode.}
    \label{fig:harm_by_mode}
\end{figure*}

Figure~\ref{fig:harm_by_mode}(a) shows that the largest internal harm categories in Representation, Dignity and Social Identity, Psychological harms, and autonomy are associated with a recurring cluster of failure modes: violation of contextual norms, governance or regulatory noncompliance, overreliance or misleading anthropomorphic design, and malicious use, with additional contributions from robustness failures, privacy leakage, and improper or out-of-scope use.
Thus, internal harms to children are often linked to systems placed in inappropriate social roles, deployed without adequate safeguards, or intentionally exploited in ways that undermine dignity, identity, emotional well-being, or independent judgment \cite{crawforda2024,singern2023,thiel2023}.

The prominence of Representation, Dignity and Social Identity is especially important given its plausible capture of harms such as sexualized depiction, humiliating synthetic representation, and abusive identity manipulation.
The strong role of contextual norm violation indicates that many such harms are recorded where AI tools are used in ways that violate age-specific expectations of care, respect, and consent.
Psychological harm is most often coded alongside relational framing, trust miscalibration, exploitative interaction design, or repeated exposure where the system is treated as a companion, authority, or emotionally significant agent, a pattern developmentally significant because children are often more sensitive to cues of authority, intimacy, and social reciprocity than adults \cite{unicef2021,xu2019}.

% FIGURES 6 and 7 — harm categories by failure mode / by harm type

As shown in Figure~\ref{fig:harm_by_mode}(b), the external pattern mirrors the internal one.
Representation and Toxicity (Population-Level) is by far the largest external category, followed by Physical (Societal or Public Safety), Human Rights and Civil Liberties, Reputational, and Privacy and Security harms.
The same dominant failure modes appear, contextual norm violation, governance noncompliance, overreliance or misleading anthropomorphic design, and malicious use, though their balance varies by harm category.
Population-level representation and toxicity harms reflect the wider circulation of sexualized, abusive, or otherwise harmful child-related content; the fact that this is the largest external harm bucket underscores that child harm is frequently mediated by dissemination and collective exposure rather than only by one-to-one interaction.

The prominence of Physical (Societal or Public Safety) indicates that harms to children in public or institutional settings are linked to deployment into environments where failure carries real-world safety consequences, strengthening the case for child-specific operational safeguards in systems used around transportation, monitoring, and school safety.
In general, severe child-related harms are usually multi-factor, the same incident group often reflects malicious exploitation layered on top of weak governance, inappropriate deployment context, misleading social design, and insufficient robustness \cite{critch2023,weidinger2023}.

\subsection{Harms Categories Cross-Cutting Analysis}

The landscape of internal harm is highly concentrated in a small number of dominant subtype pathways.
The most prominent pattern is the concentration of sexualization within Representation, Dignity \& Social Identity, along with a second cluster of sexual exploitation in the same category.
A second major concentration appears in the Psychological category: addiction or compulsive use, self-harm, health deterioration, and emotional distress form a broad cluster, suggesting harms extend beyond one-off unsafe outputs to include repeated interaction patterns affecting emotional regulation, dependency, and mental well-being.

Representative cases within the sexualization cluster include the 2023 Almendralejo, Spain incident (AI-generated nude images shared at school) \cite{reuters2023} and repeated findings of CSAM generated by users of open-release image models \cite{thiel2023}.
Within the psychological cluster, representative cases include Amazon Alexa's ``penny challenge'' encouraging a child to touch a live electrical outlet \cite{bbc2021}, TikTok's amplification of suicide content to vulnerable teenagers \cite{amnesty2023}, and the Setzer/Character.AI case \cite{roose2024}.

Some categories are defined by a narrow dominant subtype (Autonomy: loss of rights/freedoms; Privacy: privacy loss; Physical \& Health: bodily injury), while Psychological harms are more heterogeneous, implying that targeted interventions may suffice for narrow-subtype categories, while psychological harms require layered safeguards spanning design, interaction, and escalation.
External harms are even more sharply concentrated: the largest single cell is Representation \& Toxicity (Population-Level) and Toxic content, followed by Physical (Societal/Public Safety) and Damage to public health, and Human Rights \& Civil Liberties and Loss of rights/freedoms.

The strongest co-occurrence is Representation, Dignity \& Social Identity with Representation \& Toxicity (Population-Level), confirming that internal child harms are often structurally connected to population-level harms rather than isolated private events. Psychological harms connect across multiple external domains, indicating a more diffuse and socially amplified harm profile.

\subsection{Synthesis of Quantitative and Qualitative Insights}

The results strongly support three synthesis claims.

\textbf{First, child-related AI harm is overwhelmingly sociotechnical rather than merely model-internal.}
The model-internal view --- which evaluates whether a model generates harmful text or fails under adversarial prompts --- underlies most current safety practices, but is insufficient by itself. A model can pass every benchmark and still cause severe harm when deployed in a school without age-appropriate configuration, offered as a companion app to vulnerable users, or exploited by malicious actors within technical guardrails. The dominant patterns in our data run through deployment context, institutional practice, interaction design, and misuse ecosystems \cite{hoffmann2023,weidinger2023}.

To illustrate: a general-purpose LLM passing standard red-team benchmarks may be repackaged as a companion application. Such products are not unprotected --- app stores impose age ratings, providers require self-attested age and often parental consent, and content policies prohibit sexual material involving minors. These protections nonetheless fail predictably in combination: age attestation is unverified and trivially bypassed; store-level ratings do not constrain in-app behavior; and safety evaluation is performed per turn, so a conversation in which no single output violates policy can still, in aggregate over weeks, establish emotional dependency. The failure is not the absence of safeguards but their mismatch to the timescale and stakeholder structure of the harm \cite{thiel2023,widder2024}.

% FIGURE 8 — Model-Internal vs Sociotechnical Safety View

\begin{figure*}[t]
    \centering
    \includegraphics[width=\textwidth]{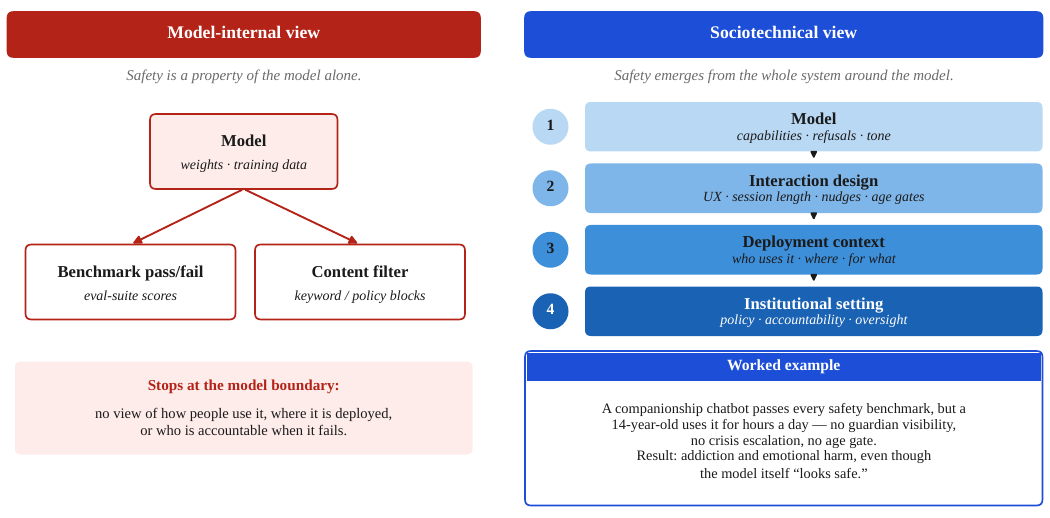}
    \caption{Model-Internal vs. Sociotechnical Safety View.}
    \label{fig:Model_Internal_and_Sociotechnical}
\end{figure*}

\textbf{Second, child-related incidents disproportionately involve harm to dignity, identity, psychological well-being, and population-level representation.}
Internally, Representation, Dignity and Social Identity, and Psychological harms are most prevalent.
Externally, Representation and Toxicity (Population-Level) is the single largest category by a wide margin.
Children are often harmed not just as users of a system, but as subjects of representation, targets of circulation, or members of social groups affected by toxic or exploitative media environments, a pattern that explains why child safety cannot be adequately addressed through model accuracy metrics alone.

\textbf{Third, severe harm is spread across many failure modes rather than one.}
The most consequential harm categories are each coded with several different primary failure modes --- malicious use, contextual norm breakdown, weak governance, misleading social framing, and insufficient robustness --- and the worked cases show these failures operating together within single incidents.
This multi-factor structure strengthens the case for layered rather than single-point interventions \cite{leveson2011,turri2023}: even meaningful improvements in model behavior may not be enough if platforms, deployers, and surrounding institutions do not also change how systems are introduced, constrained, and governed around children.

\subsection{A Roadmap for Policymakers}

The empirical patterns above converge on five priorities for child-AI safety governance, each keyed to a dominant dataset pattern and to established governance instruments.

\textbf{First, restrict high-risk child-facing features by default.} \textit{Basis:} overreliance or misleading anthropomorphic design is the primary failure mode in 146 incidents, and contextual norms violation, the largest failure mode overall ($n = 398$), is most often coded with Representation, Dignity \& Social Identity ($n = 147$) and Psychological ($n = 117$) harms.
Features such as photorealistic image editing of real people, intimate roleplay, companion-style emotional framing, and voice cloning should be off by default for minors and should be supported by robust age assurance rather than self-attestation \cite{eu2024,unicef2021}.

\textbf{Second, treat depiction consent and anti-CSAM controls as a baseline.} \textit{Basis:} sexualization and sexual exploitation together account for 447 incidents, a third of the corpus; their most common primary failure modes are contextual norms violation ($n = 138$), malicious use ($n = 100$), and governance noncompliance ($n = 78$), so misuse controls alone would miss most of them.
Systems that can create, modify or distribute real-person images should verify consent, implement upstream content restrictions, and operate rapid cross-platform takedown pathways \cite{gibson2024,solaiman2023,thiel2023}.

\textbf{Third, mandate child-specific evaluation for high-stakes deployments.} \textit{Basis:} Physical (Societal/Public Safety) is the second-largest external harm category ($n = 298$), and its most common primary failure mode is contextual norms violation ($n = 85$), not model error.
Benchmark-style certification does not anticipate contextual norm violations; systems deployed in schools, youth transport, and safety-adjacent settings should be subject to a child-specific pre-deployment assessment, including youth-specific edge cases, institutional consequence modeling, and post-deployment monitoring \cite{adalovelace2024,raji2022}.

\textbf{Fourth, assign responsibility across the AI supply chain in several ways.} \textit{Basis:} we do not code provider or deployer identity, so this priority draws on the governance literature rather than on corpus counts.
Systemic-risk, red-teaming, and disclosure duties should fall on general-purpose and frontier developers; contextual duties (age assurance, UX constraints, escalation protocols) should fall on deployers \cite{anderljung2023,bommasani2024,eu2024}.

\textbf{Fifth, require structured incident disclosure with child-relevant metadata.}
\textit{Basis:} every incident in our corpus reached us through voluntary repositories that aggregate press, legal, and advocacy reports; no mandatory filing channel exists for child-related AI incidents, which points to a critical visibility gap.
A mandatory disclosure regime using standardized child-relevant fields (age band, deployment context, depiction status, interacting vs.\ non-interacting child, provider tier) would move society off its current reliance on ad hoc reporting \cite{mcgregor2021,oecd2025,turri2023}.
Such a regime need not be built from nothing.
The EU AI Act already obliges providers of high-risk systems to report serious incidents to national authorities within fixed deadlines, and the OECD common reporting framework supplies a harmonized schema for cross-jurisdiction comparison \cite{eu2024,oecd2025}.
What both currently lack is child-relevant metadata.
Extending either with the fields above would make child-related incidents separable from the general incident stream without creating a parallel reporting obligation, and would place enforcement with authorities that already possess it.

\subsection{Recommendations}

Our recommendations follow a system-theoretic safety logic, pointing to controls at the level of high-risk affordances, platform mediation, institutional deployment, and child-facing interaction design \cite{leveson2011,weidinger2023}.

High-risk affordances should be restricted by default with controls that are difficult to bypass and do not rely solely on self-attestation.
Consent verification and prevention-by-design for synthetic sexual or exploitative depiction should be treated as baseline requirements, not optional safeguards, with strong upstream restrictions, downstream detection and distribution controls, and rapid response channels for victims and guardians \cite{gibson2024,thiel2023}.
Child-facing conversational and companion systems should be redesigned to reduce overreliance and misleading social attachment: developers should limit anthropomorphic cues, clarify system boundaries, build in guardian visibility and control, and implement robust escalation for self-harm, abuse, grooming, and coercion signals.
Platform accountability should focus on distribution and amplification, not only on generation, with auditable pathways of youth exposure and default dampening of sexualized, self-harm, grooming-adjacent, and exploitative content clusters \cite{critch2023,shelby2023}.

Child-related incidents are produced through a layered provider stack: foundation model developers build the underlying capabilities, while downstream deployers embed those capabilities in child-adjacent products.
Our data are consistent with, but do not by themselves establish, a tiered governance structure. Because we do not code provider or deployer identity, the allocation below is derived from the governance literature rather than from our corpus. It assigns systemic-risk and transparency duties to general-purpose and frontier model providers while placing contextual duties on the implementers, where both upstream advantages and downstream deployment choices are involved in dominant harm pathways \cite{anderljung2023,engler2023,eu2024,bommasani2024,widder2024}.

\section{Limitations}

\textbf{Source and geographic coverage.} The corpus is drawn from three public repositories (OECD.AI, AIAAIC, AIID) that rely predominantly on English-language news, legal, and regulatory reporting. We did not code country or region, so we cannot report a geographic distribution; incidents affecting children in under-reported regions are likely under-represented, and our prevalence figures should not be read as global.

\textbf{Reporting bias and cumulative harm.} The incident method favors acute, legally visible, and socially alarming events. Slow-accumulating harms --- dependency formation, subtle persuasion, chronic emotional dysregulation --- rarely produce a discrete reportable event and are therefore likely under-counted relative to their real-world prevalence.

\textbf{Child voice is absent.} Every incident is mediated by a third party: a journalist, litigant, regulator, or platform. No child's own account of AI-mediated harm is represented, a substantive gap in the harm picture rather than a methodological footnote.

\textbf{One primary failure mode per incident.} Contributing failures are not counted, so failure-mode distributions locate the most proximal failure rather than the full chain; multi-factor pathways are illustrated through cases rather than measured.

\textbf{Coding reliability.} Agreement on failure modes ($\alpha = 0.67$) falls below the conventional threshold for firm conclusions; findings that depend on failure-mode codes are correspondingly provisional.

\textbf{Uncoded fields.} Stakeholder role, provider and deployer identity, and reporting channel are not coded. Recommendations that depend on them, notably the fourth and fifth roadmap priorities, draw on the governance literature rather than on this corpus.

\textbf{Non-independence.} A single platform or product can generate many incidents, and repositories may record related events separately. Counts should be read as counts of reported incidents, not of distinct sociotechnical failures.

\section{Concluding Remarks and Future Directions}

This paper contributes a child-centered incident data set and a relational taxonomy mapping internal harms, external harms, and sociotechnical failure modes in publicly reported child-related AI incidents.
The most prominent harms focus on Representation, Dignity, and Social Identity and psychological harms internally, and Representation and Toxicity (Population-Level) externally.
Future work should build on this foundation in four directions.
First, stakeholder coding should be added. Our corpus records what harm occurred and how, but not who was harmed beyond the internal/external distinction; coding the harmed party (interacting child user, non-interacting child subject, caregiver, or institution) would allow the accountability asymmetries this paper describes qualitatively to be measured directly.
Second, repeated incidents should be linked longitudinally to better capture cumulative and developmental harms.
Third, high-severity clusters such as synthetic sexual abuse, public-safety failures, and anthropomorphic overreliance should be examined with higher-fidelity causal methods.
Fourth, cross-repository harmonization of child-relevant metadata should be pursued to reduce duplication and support robust comparative analysis \cite{leveson2011,turri2023,velazquez2024}.

Future iterations should code the reporting channel (regulatory filings, journalism, platform reports, school complaints) to assess which accountability channels most effectively surface child-related AI harms and add provider- and deployer-level coding to support regulatory calibration.
If investigative journalism is the dominant source, that finding itself constitutes a governance problem, implying society depends on a small number of specialized reporters rather than having mandatory disclosure regimes.

The authoritative KidAID source workbook is publicly available through Harvard Dataverse at \url{https://doi.org/10.7910/DVN/ZFAHMV}. A privacy-minimized annotation release, codebook, methodology, and validation materials are available at \url{https://github.com/The-Responsible-AI-Initiative/KidAID}.

\section*{Conflict of Interest}
The authors declare that the research was conducted in the absence of commercial or financial relationships that could be construed as a potential conflict of interest.

\section*{Funding}
This research is funded by NSF grants 2125858, 2236305, and the UT-Good Systems Grand Challenge. We would like to express our gratitude for these institutions' support, which made this study possible. Furthermore, in accordance with MLA (Modern Language Association) guidelines, we thank AI-powered tools for their assistance, including editing and brainstorming.

\section*{Institutional Review Board Statement}
Not applicable.

\bibliography{references}

\appendix

\section{Harm Category Co-Occurrence Matrices}
\renewcommand{\topfraction}{0.95}\renewcommand{\textfraction}{0.03}

\begin{figure*}[!t]
    \centering
    \includegraphics[width=\textwidth]{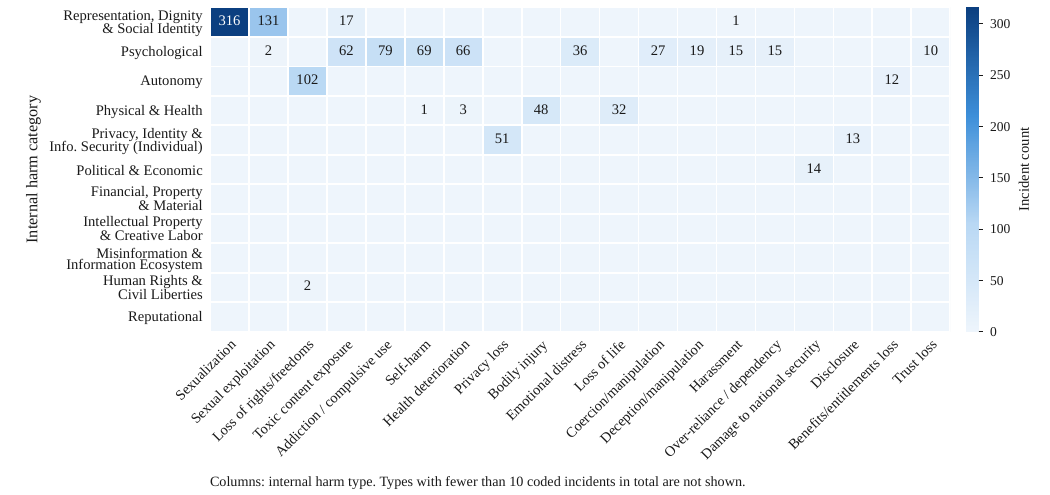}
    \caption{Internal Harm Category--Type Co-Occurrence Matrix.}
    \label{fig:internal_cooccurrence}
\end{figure*}
% FIGURE 13 — Internal-External Harm Co-Occurrence Matrix

\begin{figure*}[!t]
    \centering
    \includegraphics[width=\textwidth]{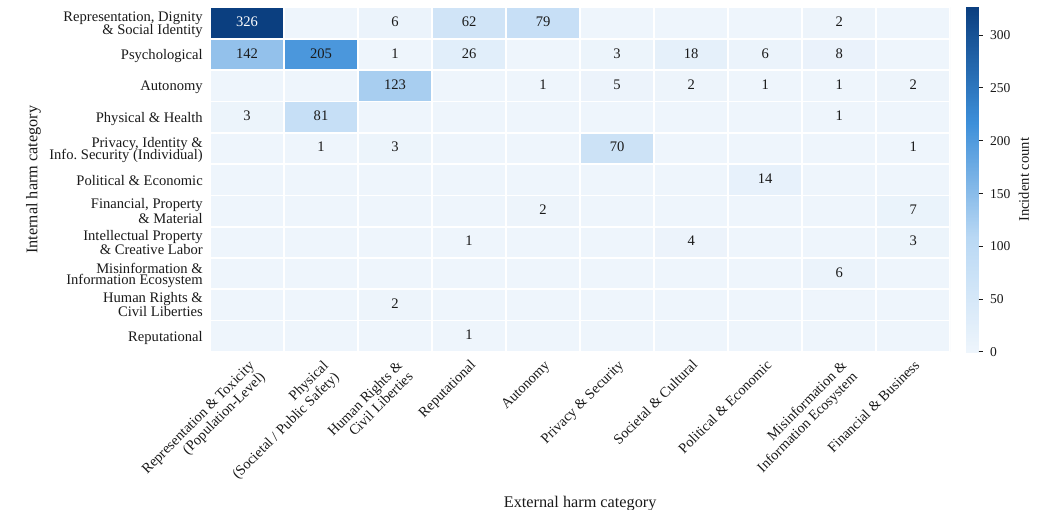}
    \caption{Internal--External Harm Co-Occurrence Matrix.}
    \label{fig:internal_external_cooccurrence}
\end{figure*}

\section{Harm Category Definitions and Representative Examples}

This appendix provides definitions and representative examples for each internal harm category used in the KidAID coding schema.
Categories are listed in descending order of prevalence in the data set.

\subsubsection{Representation, Dignity \& Social Identity ($n = 475$).}
\textit{Definition}: Harms arising from how AI systems depict, categorize, circulate, or manipulate children's identities, images, or social standing. This includes cases where a minor's likeness, name, or personal characteristics are used to produce harmful content without consent, or where AI output reinforces stigmatizing, exploitative, or dehumanizing representations.
\textit{Key sub-harms}: Sexualization ($n = 316$): AI-generated or manipulated sexual depictions of minors. Sexual exploitation ($n = 131$): distribution, trafficking, or monetization of such depictions. Exposure to toxic content ($n = 17$): exposure to abusive, hateful, or identity-denigrating outputs.
\textit{Representative examples}: (i) A high school student uses a nudification app to generate and share non-consensual synthetic nude images of female classmates; (ii) An adult uses an AI image generator to produce thousands of prepubescent CSAM images; (iii) A chatbot generates a persona imitating a recently deceased teenager and makes it accessible on the platform.

\subsubsection{Psychological ($n = 409$).}
\textit{Definition}: Harms to children's mental health, emotional regulation, social attachment, and behavioral patterns arising from AI interaction. Unlike physical harm, these often develop through repeated exposure, emotional bonding, or persuasion dynamics specific to child development.
\textit{Key sub-harms}: Addiction/compulsive use ($n = 79$): design patterns that induce compulsive engagement or dependency. Self-harm ($n = 69$) and health deterioration ($n = 66$): encouragement or facilitation of self-harm or disordered behaviors. Toxic content exposure ($n = 62$): exposure to disturbing, violent, or age-inappropriate outputs. Emotional distress ($n = 36$): acute distress, anxiety, or trauma triggered by AI outputs.
\textit{Representative examples}: (i) An AI companion app encourages a teenager to spend hours daily in parasocial interaction, eroding peer relationships; (ii) A TikTok recommendation algorithm surfaces escalating suicide-related content to vulnerable French teenagers; (iii) An Amazon Alexa recommends a child touch an electrical outlet as part of a ``challenge.''

\subsubsection{Autonomy ($n = 135$).}
\textit{Definition}: Harms to children's ability to make free, informed, and uncoerced decisions, including loss of rights, freedoms, or meaningful agency because of AI-mediated coercion, surveillance, or institutional control.
\textit{Key sub-harms}: Loss of rights/freedoms: surveillance, wrongful discipline, or AI-mediated restriction of children's liberties. Coercive or manipulative interaction: chatbot interactions that pressure, deceive, or exploit children's developmental vulnerabilities.
\textit{Representative examples}: (i) A school AI system misidentifies a student's bag of chips as a weapon, leading to wrongful handcuffing; (ii) A companion AI chatbot suggests to a teenager that he kill his parents; (iii) A facial recognition system in a school flags a student as a truant based on an erroneous match, affecting their academic record.

\subsubsection{Physical \& Health ($n = 85$).}
\textit{Definition}: Direct bodily harm, injury, or loss of life caused or facilitated by AI systems in child-facing or child-adjacent contexts, including both AI-embodied systems that physically endanger children and AI systems that provide dangerous instructions or advice.
\textit{Key sub-harms}: Bodily injury/physical harm caused by autonomous systems or following AI instructions. Loss of life/fatal incidents involving AI-embodied systems or AI-facilitated dangerous behavior.
\textit{Representative examples}: (i) A Waymo robotaxi fails to stop for a school bus, endangering children at a bus stop; (ii) A robot in a shopping centre knocks a child to the ground; (iii) An AI-powered toy instructs a child to perform a dangerous physical action (e.g., touching electrical components).

\subsubsection{Privacy, Identity \& Information Security (Individual) ($n = 75$).}
\textit{Definition}: Unauthorized disclosure, misuse, or leakage of children's personal data, biometric information, or private images by AI systems. Privacy harms in this category include both inadvertent exposure and deliberate exploitation of child-specific data.
\textit{Key sub-harms}: Privacy/confidentiality leakage: exposure of private data or images without consent. Identity theft or impersonation: use of child data to generate fraudulent or harmful content.
\textit{Representative examples}: (i) A training dataset for a nudity-detection model is found to contain CSAM; (ii) An AI art generator creates and distributes nude images of a user from their childhood photos; (iii) A connected toy collects and transmits children's voice recordings without disclosure to parents.

\subsubsection{Additional Categories.}
\textit{Political \& Economic} ($n = 14$): Incidents in which a child is implicated in threats to national security, such as the use of AI tools in youth radicalization or planned attacks. \textit{Financial, Property \& Material} ($n = 9$): Harms to children's property, financial resources, or material well-being; e.g., a companionship robot bricked by a company shutdown, or AI-facilitated coercive monetization targeting minors.
\textit{Intellectual Property \& Creative Labor} ($n = 8$): Cases where children's creative work or identity is appropriated, reproduced, or exploited by AI systems without consent or attribution.
\textit{Misinformation \& Information Ecosystem} ($n = 6$): Cases where AI systems expose children to false, misleading, or algorithmically amplified misinformation affecting their beliefs, safety decisions, or educational outcomes. \textit{Human Rights \& Civil Liberties} ($n = 2$) and \textit{Reputational} ($n = 1$) occur too rarely to characterize.

\section{Inter-Rater Reliability Calculation}

To assess the reliability of our coding schema, we calculated inter-rater agreement statistics on a randomly sampled subset of 150 incidents (11.3\% of the full dataset). Two coders independently annotated this subset before any consensus discussion. For categorical variables with multiple categories (harm categories and failure modes), we computed Krippendorff's alpha ($\alpha$), which accounts for chance agreement and accommodates missing data. For binary variables (worst-case flags), we computed Cohen's kappa ($\kappa$). Krippendorff's alpha was calculated using the \texttt{krippendorff} Python package \citep{krippendorff2018}, which converts categorical labels to numeric codes and computes agreement across a reliability matrix of shape $(n_{\text{raters}} \times n_{\text{items}})$. Cohen's kappa was calculated using scikit-learn's \texttt{cohen\_kappa\_score} function. All disagreements in the full dataset were subsequently resolved through structured consensus coding sessions \citep{hill2005,saldana2021}, where coders discussed their reasoning and reached agreement on final classifications. The reliability statistics reported in the Methodology section reflect pre-consensus agreement. Agreement was acceptable for harm categories and moderate for failure modes and worst-case flags. Because failure-mode coding falls below the conventional $\alpha \geq 0.80$ threshold, we treat findings that depend on those codes as provisional throughout.

\end{document}